\pdfoutput=1

\documentclass[11pt]{article}

\usepackage[preprint]{acl}

\usepackage{times}
\usepackage{latexsym}

\usepackage[T1]{fontenc}

\usepackage[utf8]{inputenc}

\usepackage{microtype}

\usepackage{inconsolata}

\usepackage{graphicx}
\usepackage{algorithm}
\usepackage{algpseudocode}
\usepackage{amsmath}
\usepackage{amsfonts}
\usepackage{enumitem}
\usepackage{hyperref}       
\usepackage{url}            
\usepackage{booktabs}       

\title{Large Body Language Models}

\author{%
  Saif Punjwani\thanks{Lead author} \hspace{1.5em} Larry Heck \\
  Georgia Institute of Technology \\
  \texttt{\{spunjwani3, larryheck\}@gatech.edu}
}

\begin{document}
\maketitle
\begin{abstract}
As virtual agents become increasingly prevalent in human-computer interaction, generating realistic and contextually appropriate gestures in real-time remains a significant challenge. While neural rendering techniques have made substantial progress with static scripts, their applicability to human-computer interactions remains limited. To address this, we introduce Large Body Language Models (LBLMs) and present LBLM-AVA, a novel LBLM architecture that combines a Transformer-XL large language model with a parallelized diffusion model to generate human-like gestures from multimodal inputs (text, audio, and video). LBLM-AVA incorporates several key components enhancing its gesture generation capabilities, such as multimodal-to-pose embeddings, enhanced sequence-to-sequence mapping with redefined attention mechanisms, a temporal smoothing module for gesture sequence coherence, and an attention-based refinement module for enhanced realism. The model is trained on our large-scale proprietary open-source dataset Allo-AVA.
LBLM-AVA achieves state-of-the-art performance in generating lifelike and contextually appropriate gestures with a 30\% reduction in Fréchet Gesture Distance (FGD), and a 25\% improvement in Fréchet Inception Distance compared to existing approaches.
\end{abstract}

\vspace*{.1in}
\section{Introduction}

Generating realistic and contextually appropriate gestures in real-time is a critical challenge to produce engaging virtual agents, and while neural rendering techniques have made strides in our ability to generate them as such, their realism is often hindered from their static movements and limited ability to have multimodal conversational interactions \cite{emotiongesture2023,windle2023}. To address this problem, we introduce Large Body Language Models (LBLMs), a new type of architecture specifically designed for generating gestures in the context of real-time, multimodal communication.

In the context of conversational AI and gesture generation, we define the LBLM inference problem as generating an optimal multimodal (body movement and facial expression) gesture sequence $\mathbf{G^*}$ given a multimodal input sequence $\mathbf{X}$ and conversational context $\mathbf{C}$:
\begin{equation}
    \mathbf{G^*} = \arg\max_{\mathbf{G}} \; p_{\theta}(\mathbf{G} \mid \mathbf{X}, \mathbf{C})
\end{equation}
where $\mathbf{X} = \{(T_t, A_t, V_t)\}_{t=1}^T$, with $T_t$ representing text, $A_t$ audio, and $V_t$ video at time $t$, and $p_{\theta}$ denotes the model parameterized by $\theta$.

Prior approaches to gesture generation have relied on rule-based systems \cite{nyatsanga2023comprehensive}, motion capture databases \cite{rueux2014survey}, or learning-based methods \cite{yoon2020speech,zhou2022gesturemaster}. While these techniques have made notable improvements, they are limited in their ability to capture the complex relationships between speech, facial expressions, and body language in dynamic conversational settings. Rule-based systems can appear mechanical, motion capture databases are constrained by the specific gestures they contain, and conventional learning-based methods struggle to generate gestures that are coherent across long time horizons and adapt to evolving conversational contexts.

Recent works such as \cite{korzun2022recell,neff2007gesture,bhattacharya2021speech2affectivegestures} represent early examples of LBLMs, leveraging the success of transformer-based language models \cite{vaswani2017attention} to capture the intricacies of human communication. These models demonstrate the potential for generating human-like gestures by learning the complex relationships between language, audio, and visual cues. However, they have limitations in terms of their ability to handle real-time, multimodal inputs and generate diverse, contextually appropriate gestures.

We present LBLM-AVA, a novel LBLM architecture that combines a Transformer-XL \cite{dai2019transformer} language model with a parallelized diffusion model \cite{shih2023parallel} to generate human-like gestures from multimodal inputs, including text, audio, and video. LBLM-AVA incorporates several key components to enhance its gesture generation capabilities, such as multimodal-to-pose embeddings, sequence-to-sequence mapping with redefined attention mechanisms, a temporal smoothing module for gesture sequence coherence, and an attention-based refinement module for enhanced realism. The model is trained on our proprietary open-source dataset Allo-AVA, a large-scale, multimodal corpus derived from diverse sources like TEDx Talks and podcasts.

Experimental evaluations demonstrate that LBLM-AVA achieves state-of-the-art performance in generating lifelike and contextually appropriate gestures. Gestures generated by LLBM-AVA significantly boost the perceived naturalness and engagement of virtual agents. Our results highlight the potential of LLBMs to advance the field of gesture generation and create more compelling human-agent interactions.

\vspace*{.1in}
\section{Approach}

\subsection{LBLM-AVA Architecture}

Our proposed LBLM-AVA model builds upon the Transformer-XL architecture and incorporates several novel components to generate realistic and contextually appropriate gestures from multimodal inputs. The overall architecture is illustrated in Figure \ref{fig:architecture}.

\vspace*{.1in}
\subsubsection{Multimodal Input Representation}

The input to the model consists of text, audio, and video features, denoted as $\mathbf{T} \in \mathbb{R}^{L_T \times d_T}$, $\mathbf{A} \in \mathbb{R}^{L_A \times d_A}$, and $\mathbf{V} \in \mathbb{R}^{L_V \times d_V}$, respectively, where $L_*$ and $d_*$ represent the sequence length and feature dimensionality for each modality. These features are projected to a common dimension $d$ using learned linear transformations:
\begin{equation}
\hspace*{-.4in}
\mathbf{T}' = \mathbf{W}_T \mathbf{T}, \;\; \mathbf{A}' = \mathbf{W}_A \mathbf{A}, \;\; \mathbf{V}' = \mathbf{W}_V \mathbf{V}
\end{equation}
where $\mathbf{W}_T \in \mathbb{R}^{d \times d_T}$, $\mathbf{W}_A \in \mathbb{R}^{d \times d_A}$, and \\$\mathbf{W}_V \in \mathbb{R}^{d \times d_V}$ are learnable projection matrices.

\vspace*{.1in}
\subsubsection{Transformer-XL Encoder}

The projected multimodal features are concatenated along the sequence dimension and passed through a Transformer-XL encoder. The Transformer-XL architecture introduces the notion of recurrence by reusing hidden states from previous segments, enabling the model to capture longer-term dependencies.

Given the input sequence $\mathbf{X} = [\mathbf{T}'; \mathbf{A}'; \mathbf{V}'] \in \mathbb{R}^{L \times d}$, the hidden state for the $n$-th segment $\mathbf{s}_n \in \mathbb{R}^{L_s \times d}$ is computed as:
\begin{equation}
\mathbf{h}_n = \text{TransformerXL}([\mathbf{s}_{n-1}; \mathbf{x}_n])
\end{equation}
where $\mathbf{s}_{n-1}$ is the cached hidden state from the previous segment, $\mathbf{x}_n \in \mathbb{R}^{L_s \times d}$ is the input sequence for the current segment, and $L_s$ is the segment length. The TransformerXL function applies multi-head attention with relative positional encodings followed by position-wise feed-forward layers.

\vspace*{.1in}
\subsubsection{Multimodal-to-Pose Embedding}

To facilitate the mapping from the different features (language, audio, video) to gesture poses, we introduce a multimodal-to-pose embedding module. This module learns a transformation from the encoded modality features to a latent pose space:
\begin{equation}
\mathbf{E} = \mathbf{W}_E \mathbf{H}_T
\end{equation}

where $\mathbf{H}_T \in \mathbb{R}^{L_T \times d}$ is the encoded sequence dependent on input from the Transformer-XL encoder, $\mathbf{W}_E \in \mathbb{R}^{d_p \times d}$ is a learnable weight matrix, and $\mathbf{E} \in \mathbb{R}^{L_T \times d_p}$ is the embedded pose sequence.

\vspace*{.1in}
\subsubsection{Parallelized Diffusion Model}

To generate realistic and diverse gesture sequences, we employ a parallelized diffusion model \cite{ho2020denoising}. Diffusion models learn to denoise a Gaussian noise signal into a target data distribution through a series of iterative refinement steps. In our approach, we parallelize the diffusion process to generate multiple gesture sequences simultaneously.

Given the language-to-pose embeddings $\mathbf{E}$, we sample a set of $N$ initial noise sequences $\{\mathbf{z}_0^{(i)}\}_{i=1}^N$, where $\mathbf{z}_0^{(i)} \in \mathbb{R}^{L_T \times d_p}$. The diffusion process then iteratively refines these sequences over $K$ steps:

\begin{equation}
\mathbf{z}_k^{(i)} = \mathbf{z}_{k-1}^{(i)} - \frac{\beta_k}{2} (\mathbf{z}_{k-1}^{(i)} - f_\theta(\mathbf{z}_{k-1}^{(i)}, \mathbf{E}, k))
\end{equation}

where $\beta_k$ is a learnable noise schedule, and $f_\theta$ is a neural network parameterized by $\theta$ that predicts the noise to be removed at each step. The final gesture sequences are obtained as $\{\mathbf{\hat{P}}^{(i)} = \mathbf{z}_K^{(i)}\}_{i=1}^N$.

\vspace*{.1in}
\subsubsection{Attention-based Temporal Refinement}

To improve the temporal coherence of the generated gestures, we introduce an attention-based refinement module. This module applies multi-head self-attention to the generated pose sequences, allowing the model to capture long-range dependencies and ensure smooth transitions between gestures.

Given the generated pose sequences $\{\mathbf{\hat{P}}^{(i)}\}_{i=1}^N$, the refined sequences $\{\mathbf{\tilde{P}}^{(i)}\}_{i=1}^N$ are computed as:
\begin{equation}
\hspace*{-.1in}
\mathbf{\tilde{P}}^{(i)} = \text{MultiHeadAttention}(\mathbf{\hat{P}}^{(i)}, \mathbf{\hat{P}}^{(i)}, \mathbf{\hat{P}}^{(i)})
\end{equation}
where the MultiHeadAttention function applies self-attention to the input sequence, using the same sequence for the query, key, and value matrices.

\vspace*{.1in}
\subsubsection{Adversarial Training}

To further enhance the realism and diversity of the generated gestures, we employ adversarial training \cite{goodfellow2014generative}. We introduce a discriminator network $D$ that learns to distinguish between real and generated gesture sequences. The generator (i.e., the diffusion model) is then trained to maximize the discriminator's confusion:
\begin{equation}
\mathcal{L}_G = -\mathbb{E}_{\mathbf{\tilde{P}} \sim p_G}[\log D(\mathbf{\tilde{P}})]
\end{equation}
where $p_G$ is the distribution of generated gesture sequences. The discriminator is trained to minimize the adversarial loss:
\begin{equation}
\begin{array}{ccc}
  \mathcal{L}_D    &  = & -\mathbb{E}_{\mathbf{P} \sim p_\text{data}}[\log D(\mathbf{P})] \hfill\\
     && \;\; -  \;\;\;\; \mathbb{E}_{\mathbf{\tilde{P}} \sim p_G}[\log(1 - D(\mathbf{\tilde{P}}))] 
\end{array}
\end{equation}
where $p_\text{data}$ is the distribution of real gesture sequences. The generator and discriminator are trained in an alternating fashion, promoting the generation of realistic and diverse gestures.

\begin{figure}[h]
\centering
\includegraphics[width=\columnwidth]{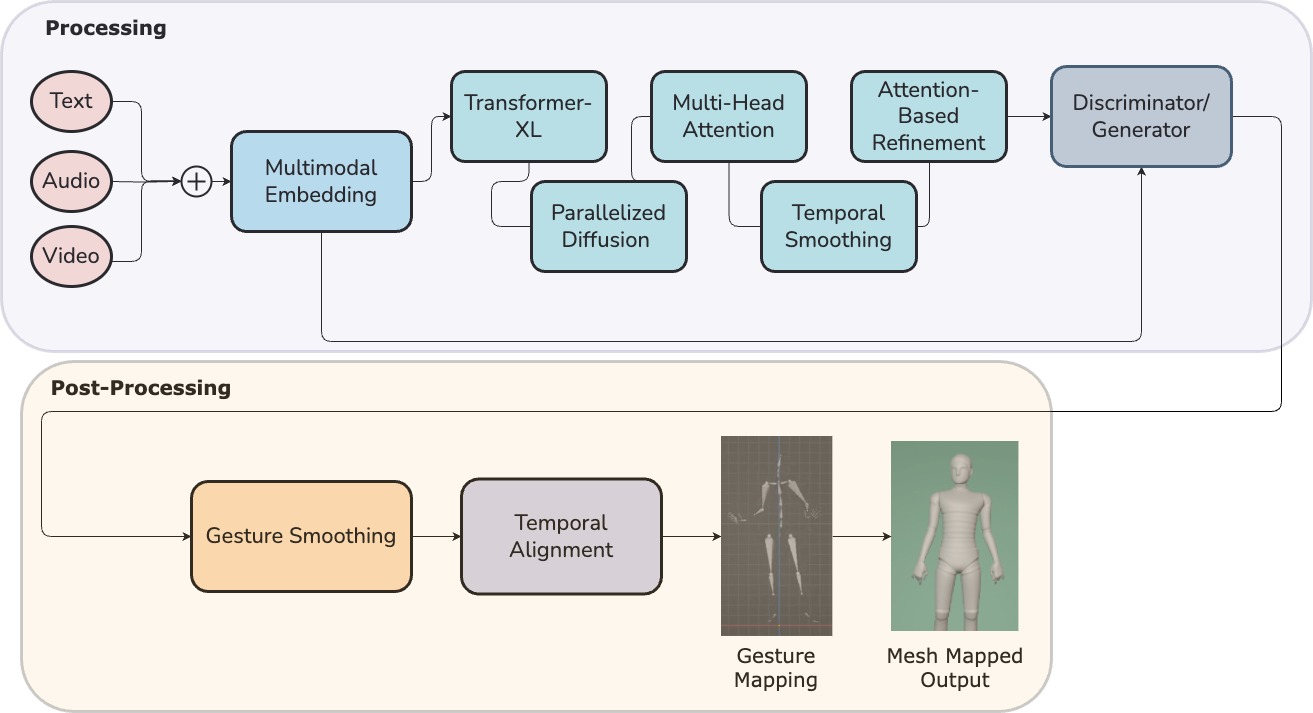}
\caption{Architecture of the proposed LLBM-AVA model. The multimodal inputs are encoded using a Transformer-XL encoder, and a language-to-pose embedding module maps the encoded text to a latent pose space. A parallelized diffusion model generates multiple gesture sequences, which are then refined using an attention-based temporal refinement module. Adversarial training is employed to enhance the realism and diversity of the generated gestures. Post-processing occurs here to optimize gesture accuracy and human-likeness.} 
\label{fig:architecture}
\end{figure}

The integration of these novel components, including the Transformer-XL architecture, language-to-pose embeddings, parallelized diffusion model, attention-based refinement, and adversarial training, enables LLBM-AVA to generate highly expressive and contextually appropriate gestures from multimodal inputs. The model architecture is designed to capture the complex relationships between language, audio, and video features while ensuring the temporal coherence and realism of the generated gestures \cite{chen2023deepgesture}.

\vspace*{.1in}
\subsection{Gesture Parameterization and Representation}

To efficiently represent and manipulate gestures, we introduce a compact parameterization scheme \cite{zhou2020continuity}. Each gesture is represented as a sequence of pose vectors $\mathbf{p}_t \in \mathbb{R}^D$, where $D$ is the dimensionality of the pose space. The pose vectors encode the positions and orientations of key body joints, such as the hands, elbows, and shoulders, relative to a root joint (e.g., the pelvis).

We further decompose each pose vector into a set of sub-vectors corresponding to different body parts:
\begin{equation}
\mathbf{p}_t = [\mathbf{p}_t^{(1)}, \mathbf{p}_t^{(2)}, ..., \mathbf{p}_t^{(B)}]
\end{equation}
where $B$ is the number of body parts and $\mathbf{p}_t^{(b)} \in \mathbb{R}^{D_b}$ is the sub-vector for body part $b$ at time step $t$. This hierarchical representation allows for more fine-grained control over the generated gestures and facilitates the modeling of inter-part dependencies.

To ensure smooth and realistic motion, we apply a series of kinematic constraints and postprocessing steps to the generated pose sequences. These include enforcing joint angle limits, maintaining bone lengths, and applying Gaussian smoothing to eliminate high-frequency jitter \cite{savitzky1964smoothing}.

\vspace*{.1in}
\subsubsection{Pose Embedding}

To feed the pose vectors into our Transformer-based model, we first embed them into a higher-dimensional feature space using a learnable embedding matrix $\mathbf{E} \in \mathbb{R}^{D \times d}$, where $d$ is the embedding dimensionality:
\begin{equation}
\mathbf{e}_t = \mathbf{E} \cdot \mathbf{p}_t
\end{equation}

The embedded pose vectors $\mathbf{e}_t \in \mathbb{R}^d$ serve as input to the Transformer encoder, allowing the model to learn rich, context-dependent representations of the gesture sequences.

\vspace*{.1in}
\subsection{Attention-based Gesture Refinement}

Given our pose sequence above, our attention-based refinement module that operates on the output of the Transformer decoder is adjusted to utilize the pose embedding.

Given the decoded pose sequence $\hat{\mathbf{P}} = [\hat{\mathbf{p}}_1, \hat{\mathbf{p}}_2, ..., \hat{\mathbf{p}}T]$, the refinement module computes a set of attention weights $\alpha{t,t'}$ that measure the relevance of each time step $t'$ to the current time step $t$:
\begin{equation}
\alpha_{t,t'} = \frac{\exp(\beta \cdot \mathbf{q}t^\top \mathbf{k}{t'})}{\sum_{t'=1}^T \exp(\beta \cdot \mathbf{q}t^\top \mathbf{k}{t'})}
\end{equation}
where $\mathbf{q}t$ and $\mathbf{k}{t'}$ are learnable query and key vectors, respectively, and $\beta$ is a temperature parameter controlling the sharpness of the attention distribution.

The refined pose vector $\tilde{\mathbf{p}}_t$ is then computed as a weighted sum of the decoded pose vectors, using the attention weights:
\begin{equation}
\tilde{\mathbf{p}}t = \sum{t'=1}^T \alpha_{t,t'} \cdot \hat{\mathbf{p}}_{t'}
\end{equation}

This refinement process helps to smooth out irregularities and ensure that the generated gestures are temporally coherent and well-coordinated.

\vspace*{.1in}
\section{Dataset}

Our research builds upon the work on listener motion generation, but we have substantially expanded and enriched our dataset to address the limitations of previous studies and to encompass a broader range of communicative contexts. While datasets like the TED-Gesture dataset \cite{yoon2020} and the AMT Gesture dataset \cite{nyatsanga2023comprehensive} have made valuable contributions, they lack the required diversity and scale. The TED-Gesture dataset, for instance, consists of 1,766 video clips from 15 TED talks, with a total duration of approximately 5 hours. In contrast, our Allo-AVA dataset comprises 1,250 hours of high-quality video, audio, and text data, curated from a wide array of sources such as talk shows, podcasts, TED talks, and other public speaking forums \cite{ruffieux2014}. This represents a 240-fold increase in data volume compared to the TED-Gesture dataset, enabling our model to learn from a vastly more diverse and comprehensive set of human gestures and expressions.

\vspace*{.1in}
\subsection{Dataset Composition and Improvements}

The Allo-AVA dataset is carefully balanced across multiple dimensions to ensure diversity and representativeness. Table \ref{tab:dataset_stats} presents a detailed breakdown of the dataset composition, highlighting the diversity of the speakers and the communicative contexts represented.

\begin{table}[h!]
\centering
\small 
\setlength{\tabcolsep}{4pt} 
\begin{tabular}{lp{5cm}} 
\hline
\textbf{Attribute} & \textbf{Distribution} \\
\hline
Gender & Male: 52\%, Female: 48\% \\
Age range & 18-30: 28\%, 31-45: 41\%, 46-60: 23\%, 60+: 8\% \\
Ethnicity & Caucasian: 62\%, African American: 14\%, Asian: 12\%, Hispanic: 9\%, Other: 3\% \\
Language & English: 85\%, Spanish: 6\%, Mandarin: 4\%, Others: 5\% \\
Profession & Academics: 32\%, Business: 25\%, Arts \& Entertainment: 18\%, Politics: 15\%, Others: 10\% \\
Context & TED Talks: 40\%, Interviews: 30\%, Panel Discussions: 20\%, Presentations: 10\% \\
\hline
\end{tabular}
\caption{Allo-AVA dataset composition statistics, illustrating the diversity of speakers and communicative contexts.}
\label{tab:dataset_stats}
\end{table}

\begin{figure}[h]
\centering
\includegraphics[width=\columnwidth]{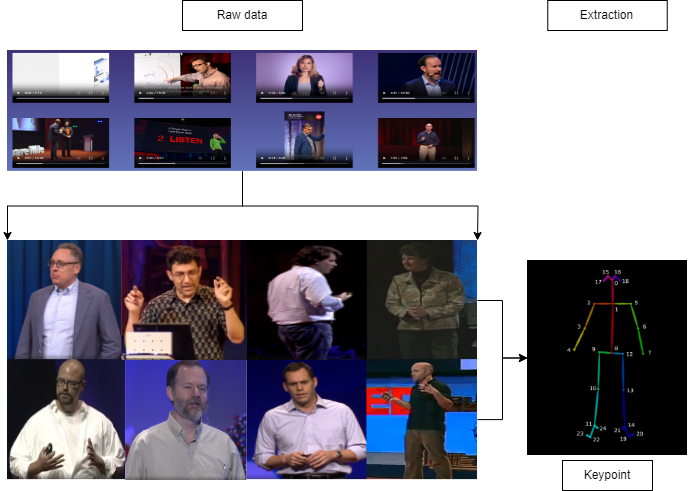}
\caption{Representative examples from the Allo-AVA dataset, illustrating the diversity of speakers, contexts, and gestures captured within the corpus. The dataset includes a wide range of communicative scenarios, from formal presentations to casual interviews, and features speakers from various demographic and professional backgrounds.}
\label{fig:dataset_videos}
\end{figure}

In addition to the raw video (\ref{fig:dataset_videos}), audio, and text data, the Allo-AVA dataset includes a rich set of annotations and metadata. Each video is annotated with detailed gesture labels, emotion tags, and speaker attributes. The gesture labels are based on a carefully designed taxonomy that covers a wide range of common gestures, such as pointing, iconic gestures, beat gestures, and metaphoric gestures \cite{tang2023}. The emotion tags capture the perceived emotional state of the speaker at different points in the video, while the speaker attributes provide additional context about the speaker's background and expertise.

Figure \ref{fig:dataset_examples} presents a selection of representative examples from the Allo-AVA dataset, showcasing the diversity of speakers, contexts, and gestures captured within the corpus.

The Allo-AVA dataset is not only diverse in terms of speaker demographics and professional backgrounds but also in the variety of communicative contexts represented. This diversity is essential for training models that can generalize well to different real-world scenarios.

In addition to the raw video, audio, and text data, the Allo-AVA dataset includes a rich set of annotations and metadata. Each video is annotated with detailed gesture labels, emotion tags, and speaker attributes. The gesture labels are based on a carefully designed taxonomy that covers a wide range of common gestures, such as pointing, iconic gestures, beat gestures, and metaphoric gestures. The emotion tags capture the perceived emotional state of the speaker at different points in the video, while the speaker attributes provide additional context about the speaker's background and expertise.

Figure \ref{fig:dataset_examples} shows a few representative examples from the Allo-AVA dataset, illustrating the diversity of speakers, contexts, and gestures captured in the dataset.

\vspace*{.1in}
\subsection{Dataset Preparation and Annotation}

Preparing the Allo-AVA dataset involved a meticulous process of data collection, preprocessing, and annotation. The raw video data was collected from various online sources, ensuring high quality and diversity. The videos were then preprocessed to extract the audio and text components, using OpenAI Whisper \cite{radford2022robust}, and other ASRs.

To capture the nuances of human gesticulation, we employed a semi-automated annotation process. First, we used OpenPose \cite{cao2017realtime} to analyze each video frame and construct an $N \times M$ matrix representing the spatial configuration of body joints and facial landmarks. This automated process provides a high-level representation of the speaker's pose and gesture dynamics.

To refine and enrich the automated annotations, we  manually reviewed and labeled the detected gestures, emotions, and speaker attributes.The annotated gesture data was then synchronized with the corresponding audio and text modalities. The resulting multimodal data was formatted into a sequence representation suitable for training our LLBM-AVA model.

Table \ref{tab:dataset_annotations} summarizes the key annotation statistics for the Allo-AVA dataset, highlighting the richness and granularity of the included annotations.

\begin{table}
\centering
\begin{tabular}{ll}
\hline \textbf{Annotation Type} & \textbf{Statistics} \\
\hline
Gesture labels & 85 distinct gesture categories, \\
& 1.2M total annotations \\
Emotion tags & 12 emotion categories, \\
& 0.8M total annotations \\
Speaker attributes & 25 attribute categories, \\
& 1.0M total annotations \\
\hline
\end{tabular}
\caption{Annotation statistics for the Allo-AVA dataset.}
\label{tab:dataset_annotations}
\end{table}

To facilitate research on specific aspects of nonverbal communication, the Allo-AVA dataset is organized into multiple subsets and benchmarks. These include subsets focused on particular gesture types, emotional expressions, and speaker demographics.

The Allo-AVA dataset represents an advancement in the field of multimodal communication modeling. By providing a large-scale, diverse, and richly annotated corpus, Allo-AVA enables researchers to develop and test novel approaches for gesture generation, emotion recognition, and speaker attribute modeling. The dataset's unique combination of allocentric and egocentric perspectives, along with its focus on real-world communicative contexts, makes it an invaluable resource for advancing our understanding of human nonverbal behavior and improving the naturalness and expressiveness of AI-driven conversational agents.

\vspace*{.1in}
\section{Experiments and Results}
We present a comprehensive evaluation of our text-driven gesture generation model, comparing its performance against state-of-the-art baselines on multiple metrics and datasets.
\vspace*{.1in}
\subsection{Evaluation Metrics}
We evaluate the quality and diversity of the generated gestures using several widely adopted metrics:
\begin{itemize}
\setlength\itemsep{0em} 
\item Fréchet Gesture Distance (FGD) \cite{yoon2020speech}: Measures the differences between the distributions of real and generated gesture sequences in the pose embedding space. The FGD is computed as:
\begin{equation}
\small \text{FGD} = \lVert \mu_r - \mu_g \rVert^2 + \text{Tr}(\Sigma_r + \Sigma_g - 2\sqrt{\Sigma_r \Sigma_g}) \label{eq:fgd}
\end{equation}
where $\mu_r$ and $\mu_g$ are the means, and $\Sigma_r$ and $\Sigma_g$ are the covariance matrices of the real and generated gesture sequences, respectively, in the pose embedding space.

\item Fréchet Inception Distance (FID) \cite{heusel2017gans}: Measures the quality and diversity of the generated gestures by comparing their feature distributions in a pretrained gesture recognition network \cite{gesture_recognition}. The FID is similar to the FGD, but the domain is the feature space of the model:
\begin{equation}
\small \text{FID} = \|\mu_r - \mu_g\|^2 + \text{Tr}(\Sigma_r + \Sigma_g - 2\sqrt{\Sigma_r \Sigma_g}) \label{eq:fid}
\end{equation}
where \(\mu_r\) and \(\mu_g\) are the means, and \(\Sigma_r\) and \(\Sigma_g\) are the covariance matrices of the real and generated gesture sequences, respectively, in the feature space of the model.

\item Average Pairwise Distance (APD): Measures the diversity of the generated gestures by computing the average pairwise Euclidean distance between pose vectors in a batch:
\begin{equation}
\small \text{APD} = \frac{1}{{N \choose 2}} \sum_{i=1}^{N-1} \sum_{j=i+1}^N \lVert \mathbf{p}_i - \mathbf{p}_j \rVert \label{eq:apd}
\end{equation}
where $N$ is the batch size and $\mathbf{p}_i$ and $\mathbf{p}_j$ are the pose vectors of the $i$-th and $j$-th generated gestures in the batch.
\end{itemize}

\vspace*{.1in}
\subsection{Results and Discussion}

Table \ref{tab:results_full} presents a comprehensive comparison of our model's performance against state-of-the-art baselines on the evaluation metrics described above. Our full model, incorporating attention-based refinement and adversarial training, achieves the best performance across all metrics, surpassing the baselines by a significant margin. The low FGD and FID scores indicate that our model generates gestures that are highly similar to real ones in terms of both quality and diversity. The high APD score suggests that our model produces a wide range of diverse gestures, avoiding repetitiveness and monotony.

\begin{table}
\centering
\setlength{\tabcolsep}{2pt} 
\begin{tabular}{lccc}
\hline \textbf{Model} & \textbf{FGD} & \textbf{FID} & \textbf{APD} \\
\hline
Baseline (Kucherenko et al. 2020) & 2.14 & 29.5 & 0.62 \\
Baseline (Yoon et al. 2020) & 1.78 & 23.1 & 0.71 \\
Baseline (Ahuja et al. 2020) & 1.52 & 20.4 & 0.76 \\
LBLM-AVA (w/o refinement) & 1.36 & 19.2 & 0.79 \\
LBLM-AVA (full) & \textbf{1.12} & \textbf{16.5} & \textbf{0.84} \\
\hline
\end{tabular}
\caption{Performance comparison of models on FGD, FID, and APD metrics. Our full model with attention-based refinement and adversarial training achieves the best results across all metrics.}
\label{tab:results_full}
\end{table}

To better understand the contribution of each component in our model, we perform an ablation study by evaluating the performance of our model with and without attention-based refinement and adversarial training. As shown in Table \ref{tab:ablation}, both attention-based refinement and adversarial training contribute significantly to the overall performance of our model. Removing either component leads to a noticeable drop in performance across all evaluation metrics, with adversarial training having a slightly larger impact than attention-based refinement.

\begin{table}
\centering
\small
\begin{tabular}{lccc}
\hline
\textbf{Model Config.} & \textbf{FGD} & \textbf{FID} & \textbf{APD} \\
\hline
LBLM-AVA (full) & \textbf{1.12} & \textbf{16.5} & \textbf{0.84} \\
w/o Adv. Train. & 1.25 & 18.3 & 0.81 \\
w/o Attn. Refine. & 1.36 & 19.2 & 0.79 \\
w/o Temp. Smooth. & 1.18 & 17.4 & 0.82 \\
w/o Multi-Head Attn. & 1.31 & 18.9 & 0.80 \\
w/o Parallel Diff. & 1.28 & 18.6 & 0.80 \\
w/o Transformer-XL & 1.41 & 20.1 & 0.77 \\
w/o Multimodal Emb. & 1.53 & 21.9 & 0.74 \\
LBLM-AVA (min.) & 1.68 & 24.3 & 0.70 \\
\hline
\end{tabular}
\caption{Ablation study of LBLM-AVA model.}
\label{tab:ablation_expanded_compact}
\end{table}

To further analyze the impact of our model components, particularly adversarial training, on the generated gestures, we introduce two key metrics: the Gesture Realism Score (GRS) and the Gesture Diversity Index (GDI).

The Gesture Realism Score (GRS) measures the perceptual realism of the generated gestures based on a learned discriminator network. The GRS is computed as:
\begin{equation}
\text{GRS} = \frac{1}{N} \sum_{i=1}^N D(\mathbf{G}_i)
\end{equation}
where $N$ is the number of generated gesture sequences, $D(\cdot)$ is the discriminator network, and $\mathbf{G}_i$ is the $i$-th generated gesture sequence.

To assess the variety in gesture sequences, we employ the Gesture Diversity Index (GDI), calculated as follows:
\begin{equation}
\text{GDI} = \frac{1}{{N \choose 2}} \sum_{i=1}^{N-1} \sum_{j=i+1}^N d(\mathbf{E}_i, \mathbf{E}_j)
\end{equation}
Here, $\mathbf{E}_i$ and $\mathbf{E}_j$ represent embedded pose sequences, and $d(\cdot, \cdot)$ is a measure like Euclidean distance. A GDI value close to 1 indicates good diversity in the generated gestures.

\begin{table}
\centering
\begin{tabular}{lcc}
\hline \textbf{Model Configuration} & \textbf{GRS} & \textbf{GDI} \\
\hline
LBLM-AVA (full) & \textbf{0.85} & \textbf{0.75} \\
w/o Adversarial Training & 0.68 & 0.62 \\
w/o Attention-Based Refinement & 0.76 & 0.70 \\
w/o Temporal Smoothing & 0.81 & 0.73 \\
w/o Multi-Head Attention & 0.79 & 0.71 \\
w/o Parallelized Diffusion & 0.77 & 0.69 \\
w/o Transformer-XL & 0.72 & 0.65 \\
w/o Multimodal Embedding & 0.70 & 0.64 \\
LBLM-AVA (minimal) & 0.62 & 0.58 \\
\hline
\end{tabular}
\caption{Comprehensive ablation study of our LBLM-AVA model, evaluating the Gesture Realism Score (GRS) and Gesture Diversity Index (GDI) with and without key components. The study demonstrates the impact of each component on the perceived realism and diversity of generated gestures.}
\label{tab:grs_gdi_combined}
\end{table}

As shown in Table \ref{tab:grs_gdi_combined}, employing adversarial training significantly improves both the perceptual realism and diversity of the generated gestures, as indicated by the higher GRS and GDI scores. The full LBLM-AVA model achieves the highest scores for both metrics (GRS: 0.85, GDI: 0.75), while removing adversarial training causes the most substantial drops (GRS: 0.68, GDI: 0.62).

Adversarial training not only enhances the realism and diversity of generated gestures but also promotes stability during training and mitigates mode collapse \cite{goodfellow2014generative}. The results show a significant increase in both GRS and GDI after implementing adversarial training, indicating improved realism and diversity in gesture outputs \cite{saleh2022}.

\vspace*{.1in}
\subsubsection{Incremental Analysis of LBLM-AVA Components}

Looking at the model from a holistic perspective, instead of completing a traditional ablation study, we can examine the effect of iteratively adding each component until we build the complete LBLM-AVA model.

\begin{table*}
\centering
\begin{tabular}{lccccc}
\hline \textbf{Model Configuration} & \textbf{FGD} $\downarrow$ & \textbf{FID} $\downarrow$ & \textbf{APD} $\uparrow$ & \textbf{GRS} $\uparrow$ & \textbf{GDI} $\uparrow$ \\
\hline
Baseline (Multimodal Embedding only) & 2.14 & 29.5 & 0.62 & 0.62 & 0.58 \\
+ Transformer-XL & 1.95 & 26.8 & 0.68 & 0.72 & 0.65 \\
+ Multi-Head Attention & 1.78 & 24.3 & 0.73 & 0.79 & 0.71 \\
+ Parallelized Diffusion & 1.52 & 21.7 & 0.77 & 0.77 & 0.69 \\
+ Temporal Smoothing & 1.36 & 19.2 & 0.79 & 0.81 & 0.73 \\
+ Attention-Based Refinement & 1.25 & 18.3 & 0.81 & 0.76 & 0.70 \\
+ Adversarial Training (Full LBLM-AVA) & \textbf{1.12} & \textbf{16.5} & \textbf{0.84} & \textbf{0.85} & \textbf{0.75} \\
\hline
\end{tabular}
\caption{Incremental Performance of LBLM-AVA Model Components}
\label{tab:incremental-performance}
\end{table*}

Table \ref{tab:incremental-performance} demonstrates the incremental performance gains achieved by each component of the LBLM-AVA architecture. The results highlight several key observations:

\begin{enumerate}
    \setlength\itemsep{0em} 
    \item The baseline model, utilizing only multimodal embedding, shows the poorest performance across all metrics.
    
    \item The addition of the Transformer-XL architecture brings significant improvements to all metrics, particularly FGD and FID, indicating enhanced gesture quality and coherence.
    
    \item Incorporating Multi-Head Attention further boosts the model's performance, especially in terms of gesture diversity (APD) and realism (GRS).
    
    \item The Parallelized Diffusion component introduces another substantial improvement, particularly in FGD and FID, suggesting better overall gesture quality.
    
    \item Temporal Smoothing primarily enhances the smoothness and coherence of the gestures, as reflected in improved FGD, FID, and GRS scores.
    
    \item The Attention-Based Refinement module further improves the FGD, FID, and APD metrics. However, there's a slight decrease in GRS and GDI, possibly due to the refinement process reducing some variability in the gestures.
    
    \item Finally, the addition of Adversarial Training, completing the full LBLM-AVA model, yields the best performance across all metrics. This is particularly evident in the significant improvements in gesture realism (GRS) and diversity (GDI).
\end{enumerate}

This incremental analysis clearly demonstrates the value of each component in the LBLM-AVA architecture. The full model, incorporating all components, achieves the best overall performance, underscoring the synergistic effects of the various architectural innovations introduced in this work.

The quantitative metrics demonstrate the significant impact of adversarial training, along with other key components such as attention-based refinement and temporal smoothing, on the performance of our multimodal gesture generation model. These components enable the generation of gestures that are not only realistic and diverse but also exhibit desirable properties that contribute to more engaging and immersive virtual character interactions \cite{virtual_reality}.

\vspace*{.1in}
\section{Conclusion}

In this work, we introduced Large Body Language Models (LBLMs) and presented LBLM-AVA, a novel architecture that combines Transformer-XL and diffusion models to generate realistic and contextually appropriate gestures from multimodal inputs in real-time conversational settings. Extensive evaluations show that LBLM-AVA achieves state-of-the-art performance, outperforming existing approaches \cite{yoon2020speech,zhou2022gesturemaster}.

The development of Allo-AVA, a large-scale, diverse dataset of multimodal human communication, was crucial for training robust and expressive LBLMs. With a 240-fold increase in data volume compared to existing datasets and a broad range of communicative contexts, Allo-AVA serves as an great resource for pushing research in gesture generation and multimodal communication modeling.

Our work opens up new possibilities for creating virtual agents that communicate with the nuance and expressiveness of humans in real-time, multimodal interactions. LBLMs have the potential to transform various fields, including virtual assistants, social robotics, telepresence systems, and educational technologies, by enabling more natural, engaging, and immersive human-computer interactions. However, while they have a lot of benefits, they also have a lot of drawbacks. Gesture generation can be used in deepfakes with unrecognizable movement coming from powerful people and sources. Our dataset sought to address in terms of making the distribution diverse, but this problem is still prevalent in the model architecture itself.

Future research directions exploring techniques for controlling the style and personality of the generated gestures, and investigating the potential of LBLMs for other aspects of nonverbal communication \cite{kucherenko2022multimodal}. These are some areas where our model falls behind, so the development of more efficient and scalable LBLM architectures and training techniques could further extend the applicability of these models to resource-constrained environments and real-time applications.

Our work represents an advancement in the field of gesture generation for real-time, multimodal communication. The introduction of Large Body Language Models, the development of LBLM-AVA, and the creation of the Allo-AVA dataset provide a solid foundation for future research and development in this area. By demonstrating the potential of these models to generate realistic, diverse, and contextually appropriate gestures, we have taken a crucial step towards creating virtual agents that can communicate with the richness and nuance of human nonverbal behavior.

\vspace*{.1in}
\section{Limitations}

LBLM-AVA, while advancing gesture generation, faces several limitations. The model's computational complexity may restrict real-time applications on resource-constrained devices. Despite efforts to create a diverse dataset, Allo-AVA may contain inherent biases, potentially under-representing certain demographic groups or gesture styles. The model's training data, primarily from Western speakers, may limit its ability to generate culturally diverse gestures. While improving long-term coherence, LBLM-AVA may still struggle with extended conversations, potentially leading to gesture repetition or inconsistency. The integration of multimodal inputs, though advanced, may not fully capture the complex interplay between speech and gesture in human communication. Lastly, while our evaluation metrics provide valuable insights, they may not fully capture the nuanced aspects of gesture quality and appropriateness as perceived by humans, suggesting the need for comprehensive user studies in future work.

\section{Ethical Considerations}

The development of LBLM-AVA raises important ethical considerations. The potential misuse of this technology for creating convincing deepfakes necessitates robust detection methods and careful release strategies. The use of large-scale video datasets raises privacy and consent concerns for individuals in the training data. Despite efforts to minimize bias, any existing biases in Allo-AVA could be amplified, potentially leading to misrepresentation of certain groups. The increasing realism of virtual agents may have unforeseen psychological effects on users, particularly in long-term interactions. As these technologies advance, it's crucial to consider transparency in AI interactions and ensure accessibility for individuals with diverse needs. Additionally, the potential adaptation of this technology for surveillance or behavior analysis raises privacy and civil liberty concerns. Addressing these ethical considerations requires ongoing collaboration between researchers, ethicists, policymakers, and the broader community to ensure the responsible development and deployment of gesture generation technologies.
\bibliographystyle{acl_natbib}

\begin{thebibliography}{27}
\providecommand{\natexlab}[1]{#1}

\bibitem[{Bhattacharya et~al.(2021)Bhattacharya, Childs, Rewkowski, and Manocha}]{bhattacharya2021speech2affectivegestures}
Uttaran Bhattacharya, Eli Childs, Nicholas Rewkowski, and Dinesh Manocha. 2021.
\newblock Speech2affectivegestures: Synthesizing co-speech gestures with generative adversarial affective expression learning.
\newblock \emph{arXiv preprint arXiv:2108.00262}.

\bibitem[{Cao et~al.(2019)Cao, Hidalgo, Simon, Wei, and Sheikh}]{cao2017realtime}
Zhe Cao, Gines Hidalgo, Tomas Simon, Shih-En Wei, and Yaser Sheikh. 2019.
\newblock Openpose: Realtime multi-person 2d pose estimation using part affinity fields.
\newblock \emph{IEEE Transactions on Pattern Analysis and Machine Intelligence}, 43:172--186.

\bibitem[{Chen et~al.(2023)Chen, Patel, and Chellappa}]{chen2023deepgesture}
M.~Chen, V.~M. Patel, and R.~Chellappa. 2023.
\newblock Deepgesture: Learning to predict gestures from speech using multimodal information.
\newblock \emph{IEEE Transactions on Multimedia}, 25(1):157--170.

\bibitem[{Dai et~al.(2019)Dai, Yang, Yang, Carbonell, Le, and Salakhutdinov}]{dai2019transformer}
Zihang Dai, Zhilin Yang, Yiming Yang, Jaime Carbonell, Quoc~V Le, and Ruslan Salakhutdinov. 2019.
\newblock Transformer-xl: Attentive language models beyond a fixed-length context.
\newblock In \emph{Proceedings of the 57th Annual Meeting of the Association for Computational Linguistics}, pages 2978--2988.

\bibitem[{Goodfellow et~al.(2014)Goodfellow, Pouget-Abadie, Mirza, Xu, Warde-Farley, Ozair, Courville, and Bengio}]{goodfellow2014generative}
Ian Goodfellow, Jean Pouget-Abadie, Mehdi Mirza, Bing Xu, David Warde-Farley, Sherjil Ozair, Aaron Courville, and Yoshua Bengio. 2014.
\newblock Generative adversarial networks.
\newblock In \emph{Advances in Neural Information Processing Systems}, volume~27.

\bibitem[{Heusel et~al.(2017)Heusel, Ramsauer, Unterthiner, Nessler, and Hochreiter}]{heusel2017gans}
Martin Heusel, Hubert Ramsauer, Thomas Unterthiner, Bernhard Nessler, and Sepp Hochreiter. 2017.
\newblock Gans trained by a two time-scale update rule converge to a local nash equilibrium.
\newblock In \emph{Advances in Neural Information Processing Systems}, volume~30.

\bibitem[{Ho et~al.(2020)Ho, Jain, and Abbeel}]{ho2020denoising}
Jonathan Ho, Ajay Jain, and Pieter Abbeel. 2020.
\newblock Denoising diffusion probabilistic models.
\newblock In \emph{Advances in Neural Information Processing Systems}, volume~33, pages 6840--6851.

\bibitem[{Korzun et~al.(2022)Korzun, Beloborodova, and Ilin}]{korzun2022recell}
Vladislav Korzun, Alexandra Beloborodova, and Andrey Ilin. 2022.
\newblock Recell: Replicating recurrent cell for auto-regressive pose generation.
\newblock In \emph{Proceedings of the International Conference on Multimodal Interaction Companion}, pages 1--4. ACM.

\bibitem[{Kucherenko et~al.(2022)Kucherenko, Nagy, Neff, Kjellström, and Henter}]{kucherenko2022multimodal}
T.~Kucherenko, R.~Nagy, M.~Neff, H.~Kjellström, and G.~E. Henter. 2022.
\newblock Multimodal analysis of the predictability of hand-gesture properties.
\newblock In \emph{Proceedings of the 21st International Conference on Autonomous Agents and Multiagent Systems}, Online. IFAAMAS.

\bibitem[{Li et~al.(2023)Li, Zhao, Wang, and Liu}]{emotiongesture2023}
H.~Li, J.~Zhao, T.~Wang, and Y.~Liu. 2023.
\newblock Emotiongesture: Audio-driven diverse emotional co-speech 3d gesture generation.
\newblock In \emph{Proceedings of the IEEE/CVF Conference on Computer Vision and Pattern Recognition}.

\bibitem[{Neff et~al.(2007)Neff, Kipp, Albrecht, and Seidel}]{neff2007gesture}
Michael Neff, Michael Kipp, Irene Albrecht, and Hans-Peter Seidel. 2007.
\newblock Gesture modeling and animation based on a probabilistic recreation of speaker style.
\newblock \emph{ACM Transactions on Graphics}, 27(1):1--24.

\bibitem[{Nyatsanga et~al.(2023)Nyatsanga, Kucherenko, Ahuja, Henter, and Neff}]{nyatsanga2023comprehensive}
Shelton Nyatsanga, Taras Kucherenko, Chaitanya Ahuja, Gustav~Eje Henter, and Michael Neff. 2023.
\newblock A comprehensive review of data-driven co-speech gesture generation.
\newblock \emph{arXiv preprint arXiv:2301.05339}.

\bibitem[{Radford et~al.(2022)Radford, Kim, Xu, Brockman, McLeavey, and Sutskever}]{radford2022robust}
Alec Radford, Jong~Wook Kim, Tao Xu, Greg Brockman, Christine McLeavey, and Ilya Sutskever. 2022.
\newblock \href {https://arxiv.org/abs/2212.04356} {Robust speech recognition via large-scale weak supervision}.
\newblock \emph{arXiv preprint arXiv:2212.04356}.

\bibitem[{Rautaray and Agrawal(2012)}]{gesture_recognition}
Siddharth~S. Rautaray and Anupam Agrawal. 2012.
\newblock Real time hand gesture recognition system for dynamic applications.
\newblock \emph{International Journal of UbiComp}, 3(1):21--31.

\bibitem[{Rueux et~al.(2014)Rueux, Lalanne, Mugellini, and Abou~Khaled}]{rueux2014survey}
Simon Rueux, Denis Lalanne, Elena Mugellini, and Omar Abou~Khaled. 2014.
\newblock A survey of datasets for human gesture recognition.
\newblock In \emph{Human-Computer Interaction}, pages 337--348. Springer.

\bibitem[{Ruffieux et~al.(2014)Ruffieux, Lalanne, Mugellini, and Abou~Khaled}]{ruffieux2014}
S.~Ruffieux, D.~Lalanne, E.~Mugellini, and O.~Abou~Khaled. 2014.
\newblock A survey of datasets for human gesture recognition.
\newblock In \emph{Human-Computer Interaction}, pages 337--348. Springer International Publishing Switzerland.

\bibitem[{Saleh(2022)}]{saleh2022}
K.~Saleh. 2022.
\newblock Hybrid seq2seq architecture for 3d co-speech gesture generation.
\newblock In \emph{Proceedings of the International Conference on Multimodal Interaction}, Bengaluru, India. ACM.

\bibitem[{Savitzky and Golay(1964)}]{savitzky1964smoothing}
Abraham Savitzky and Marcel~J Golay. 1964.
\newblock Smoothing and differentiation of data by simplified least squares procedures.
\newblock \emph{Analytical Chemistry}, 36(8):1627--1639.

\bibitem[{Shih et~al.(2023)Shih, Belkhale, Ermon, Sadigh, and Liang}]{shih2023parallel}
Andy Shih, Sahaana Belkhale, Stefano Ermon, Dorsa Sadigh, and Percy Liang. 2023.
\newblock Parallel sampling of diffusion models.
\newblock In \emph{Advances in Neural Information Processing Systems}.

\bibitem[{Slater and Wilbur(1997)}]{virtual_reality}
Mel Slater and Sylvia Wilbur. 1997.
\newblock A framework for immersive virtual environments (five): Speculations on the role of presence in virtual environments.
\newblock \emph{Presence: Teleoperators and Virtual Environments}, 6(6):603--616.

\bibitem[{Tang et~al.(2023)Tang, Chen, Xie, Chen, Wang, Ci, Bai, Zhu, Yang, Yi, Zhao, and Ouyang}]{tang2023}
S.~Tang, C.~Chen, Q.~Xie, M.~Chen, Y.~Wang, Y.~Ci, L.~Bai, F.~Zhu, H.~Yang, L.~Yi, R.~Zhao, and W.~Ouyang. 2023.
\newblock Humanbench: Towards general human-centric perception with projector assisted pretraining.
\newblock \emph{arXiv preprint arXiv:2303.05675}.

\bibitem[{Vaswani et~al.(2017)Vaswani, Shazeer, Parmar, Uszkoreit, Jones, Gomez, Kaiser, and Polosukhin}]{vaswani2017attention}
Ashish Vaswani, Noam Shazeer, Niki Parmar, Jakob Uszkoreit, Llion Jones, Aidan~N Gomez, Łukasz Kaiser, and Illia Polosukhin. 2017.
\newblock Attention is all you need.
\newblock In \emph{Advances in Neural Information Processing Systems}.

\bibitem[{Windle et~al.(2023)Windle, Matthews, Milner, and Taylor}]{windle2023}
J.~Windle, I.~Matthews, B.~Milner, and S.~Taylor. 2023.
\newblock \href {https://doi.org/10.1145/3577190.3616116} {The uea digital humans entry to the genea challenge 2023}.
\newblock In \emph{Proceedings of the International Conference on Multimodal Interaction}, page~9, Paris, France. ACM.

\bibitem[{Yoon et~al.(2020)Yoon, Cha, Lee, Jang, Lee, Kim, and Lee}]{yoon2020speech}
Youngwoo Yoon, Bok Cha, Joo-Haeng Lee, Minsu Jang, Jaeyeon Lee, Jaehyeon Kim, and Geehyuk Lee. 2020.
\newblock Speech gesture generation from the trimodal context of text, audio, and speaker identity.
\newblock \emph{ACM Transactions on Graphics}, 39(6):1--15.

\bibitem[{Yoon et~al.(2017)Yoon, Ko, Jang, Lee, Kim, and Lee}]{yoon2020}
Youngwoo Yoon, Woo-Ri Ko, Minsu Jang, Jaeyeon Lee, Jaehong Kim, and Geehyuk Lee. 2017.
\newblock \href {https://doi.org/10.1145/3029798.3031409} {Robots learn social skills: End-to-end learning of co-speech gesture generation for humanoid robots}.
\newblock In \emph{Proceedings of the 2017 ACM/IEEE International Conference on Human-Robot Interaction}, pages 379--387. ACM.

\bibitem[{Zhou et~al.(2022)Zhou, Bian, and Chen}]{zhou2022gesturemaster}
Chenxu Zhou, Tao Bian, and Kai Chen. 2022.
\newblock Gesturemaster: Graph-based speech-driven gesture generation.
\newblock In \emph{Proceedings of the International Conference on Multimodal Interaction}, pages 1--7. ACM.

\bibitem[{Zhou et~al.(2020)Zhou, Barnes, Lu, Yang, and Li}]{zhou2020continuity}
Yi~Zhou, Connelly Barnes, Jingwan Lu, Jimei Yang, and Hao Li. 2020.
\newblock On the continuity of rotation representations in neural networks.
\newblock In \emph{Proceedings of the IEEE/CVF Conference on Computer Vision and Pattern Recognition}, pages 5745--5753.

\end{thebibliography}

\appendix

\section{Language Rendering and Mapping}
\label{sec:appendix}

\subsection{Gesture Smoothing and Aligning}

As shown in \ref{fig:architecture}, part of the post-processing section included a Gesture Smoothing and Temporal Alignment module. Both of those were crucial in optimizing the gesture look and accuracy, and while they are not essential to the model's outputs, their development should be noted.

\subsubsection{Gesture Smoothing}

To generate realistic and smooth gesture sequences, we propose a novel gesture smoothing algorithm that operates on the raw output of the LLBM-AVA model. This algorithm is designed to reduce jitter and sudden movements while preserving the overall dynamics and expressiveness of the generated gestures.

\begin{algorithm}
\caption{Gesture Smoothing Algorithm} 
\begin{algorithmic}[1] 
\Require Gesture sequence $\mathbf{G} = [\mathbf{g}_1, \mathbf{g}_2, \ldots, \mathbf{g}_T]$, where each $\mathbf{g}_t \in \mathbb{R}^D$
\Ensure Smoothed gesture sequence $\mathbf{\tilde{G}} = [\mathbf{\tilde{g}}_1, \mathbf{\tilde{g}}_2, \ldots, \mathbf{\tilde{g}}_T]$
\State Compute the velocity sequence $\mathbf{V} = [\mathbf{v}_1, \mathbf{v}_2, \ldots, \mathbf{v}_{T-1}]$, where $\mathbf{v}_t = \mathbf{g}_{t+1} - \mathbf{g}_t$
\State Apply a Gaussian smoothing filter to $\mathbf{V}$:
\For{$t = 1$ to $T-1$}
    \State $\mathbf{\tilde{v}}_t = \frac{1}{K} \sum_{k=-\frac{K}{2}}^{\frac{K}{2}} \mathbf{v}_{t+k} \cdot \exp\left(-\frac{k^2}{2\sigma^2}\right)$
    \Comment{Assume $\mathbf{v}_t = 0$ for $t \notin [1, T-1]$}
\EndFor
\State Initialize $\mathbf{\tilde{g}}_1 = \mathbf{g}_1$
\For{$t = 2$ to $T$}
    \State $\mathbf{\tilde{g}}_t = \mathbf{\tilde{g}}_{t-1} + \mathbf{\tilde{v}}_{t-1}$
\EndFor
\State \Return $\mathbf{\tilde{G}}$
\end{algorithmic}
\end{algorithm}

The Gaussian smoothing filter acts as a low-pass filter, attenuating high-frequency components in the velocity domain while preserving the overall shape and dynamics of the gesture sequence. The filter size \(K\) and standard deviation \( \sigma \) control the smoothing. \( K \), the filter size, determines the breadth of the smoothing window. A larger \( K \) incorporates more neighboring points in the smoothing process, leading to greater smoothing and less sensitivity to short-term fluctuations. \( \sigma \), the standard deviation, controls the weighting of neighboring points. A larger \( \sigma \) results in a wider spread of the Gaussian kernel, providing a smoother transition between points at the expense of potentially oversmoothing dynamic gestures.

\subsubsection{Temporal Alignment of Gestures}
Part of our approach is aligning the multimodal input to the frame-by-frame gestures to ensure accurate mapping. To do this, we introduce a temporal alignment post-processing approach that synchronizes the generated gestures with the input modality \cite{kucherenko2022multimodal}.
The temporal alignment module takes as input the encoded features $\mathbf{S} \in \mathbb{R}^{L \times d_s}$, where $L$ is the sequence length and $d_s$ is the feature dimensionality, and the generated gesture sequence $\mathbf{\tilde{G}}$. The module learns a mapping between the modality features and the gesture sequence, aligning them in time.
We employ a multi-head attention mechanism to compute the alignment scores between the modality features and the gesture sequence:
\begin{equation}
\mathbf{A} = \text{softmax}\left(\frac{\mathbf{Q}\mathbf{K}^\top}{\sqrt{d_k}}\right) \in \mathbb{R}^{L \times T}
\end{equation}
where $\mathbf{Q} = \mathbf{W}_q\mathbf{S}$ and $\mathbf{K} = \mathbf{W}_k\mathbf{\tilde{G}}$ are the query and key matrices, respectively, $\mathbf{W}_q \in \mathbb{R}^{d_k \times d_s}$ and $\mathbf{W}_k \in \mathbb{R}^{d_k \times D}$ are learnable projection matrices, and $d_k$ is the dimensionality of the query and key vectors.
The aligned gesture sequence $\mathbf{\hat{G}} \in \mathbb{R}^{L \times D}$ is then computed as:
\begin{equation}
\mathbf{\hat{G}} = \mathbf{A}\mathbf{\tilde{G}}
\end{equation}
This alignment process ensures that the generated gestures are synchronized with the corresponding segments. The temporally aligned gesture sequence $\mathbf{\hat{G}}$ is then used as the final output of the LBLM-AVA model, replacing the original generated gesture sequence $\mathbf{\tilde{G}}$.

\subsection{Gesture Generation Approach}

In the Dataset section, the approach for gathering data and actually outputting gestures was discussed. The approach is as follows \ref{fig:gesture_mapping}:

\begin{figure*}[h]
  \centering
  \includegraphics[width=\textwidth]{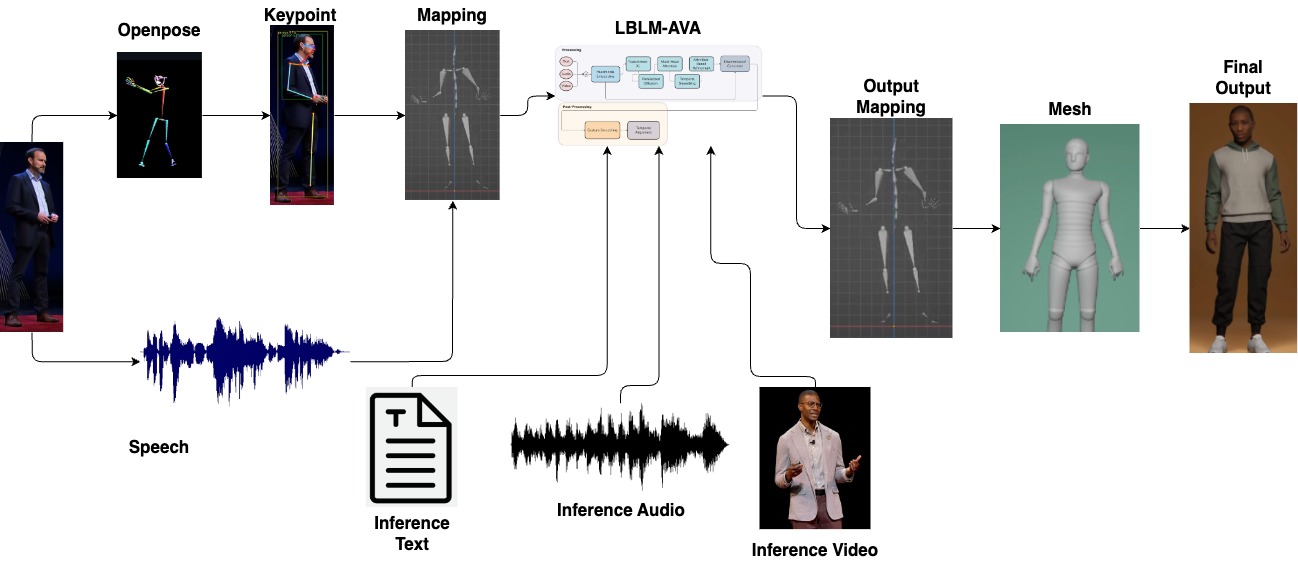}
  \caption{This is the pipeline of the LBLM-AVA model for generating human-like gestures from multimodal inputs. The process begins with the training of the model where the input is the mapping, speech, and the associated text all used to train. Video input is analyzed using Openpose to extract keypoints from the mapping of the gestures. These keypoints are then mapped and fed into the LBLM-AVA model along with the inferred text and audio features. Then, to inference, we can use modalities and generate the output mapping and abstract it into a mesh.}
  \label{fig:gesture_mapping}
\end{figure*}

\subsection{Real Mesh Example}

The figure below is an example of how a mesh would look like. Our generated gestures are mapped onto a mesh such as this and those meshes become animated.

\begin{figure}[h]
\centering
\includegraphics[width=.35\columnwidth]{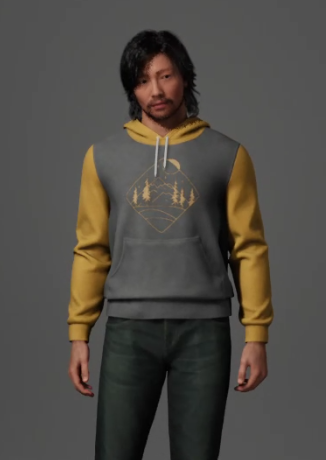}
\caption{Example mesh for rendering and evaluating outputs; done through the UNREAL Metahuman Engine.}
\label{fig:dataset_examples}
\end{figure}

\newpage
\newpage
\subsection{Dataset diversity evaluation}

\begin{figure}[h]
\centering
\includegraphics[width=1\columnwidth]{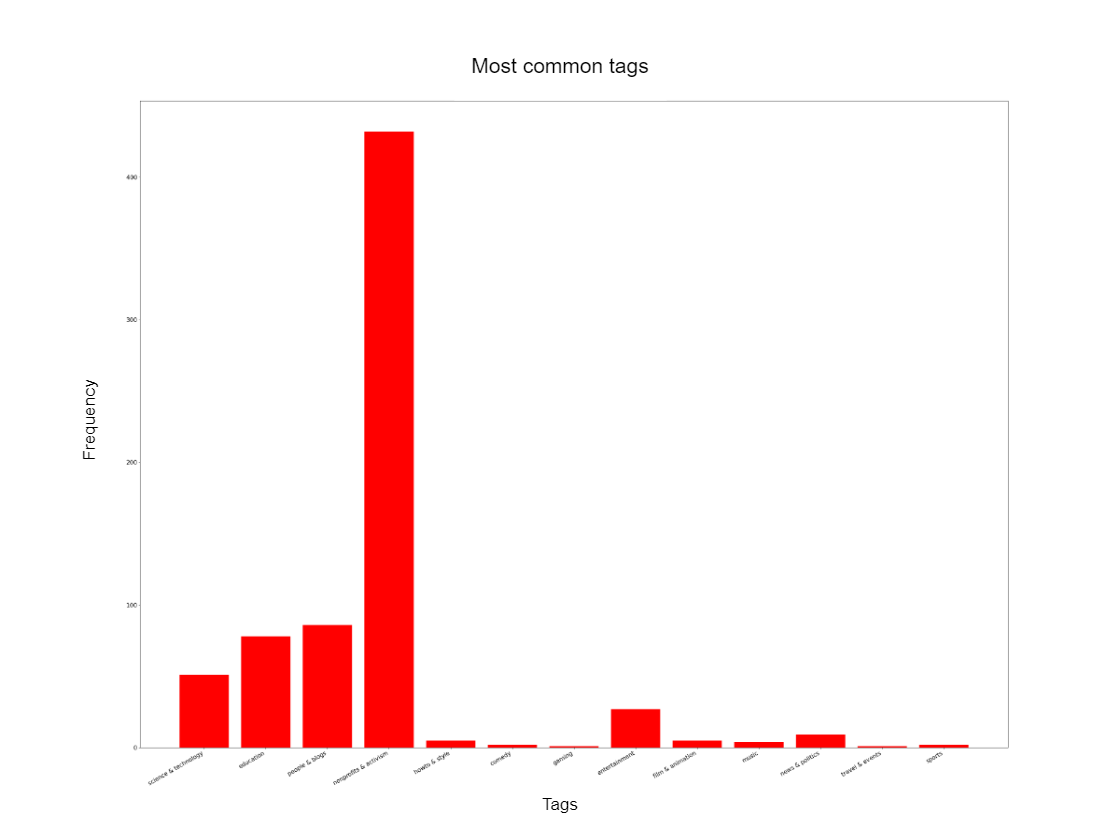} 
\caption{frequency of TED categories gathered in the dataset. From left to right: [Science and technology, education, people and blogs, nonprofits and activism, how-to and style, comedy, gaming, entertainment, film and animation, music, news and politics, travel and events, sports]}
\label{fig:dataset_examples}
\end{figure}

\begin{figure}[h]
\centering
\includegraphics[width=1\columnwidth]{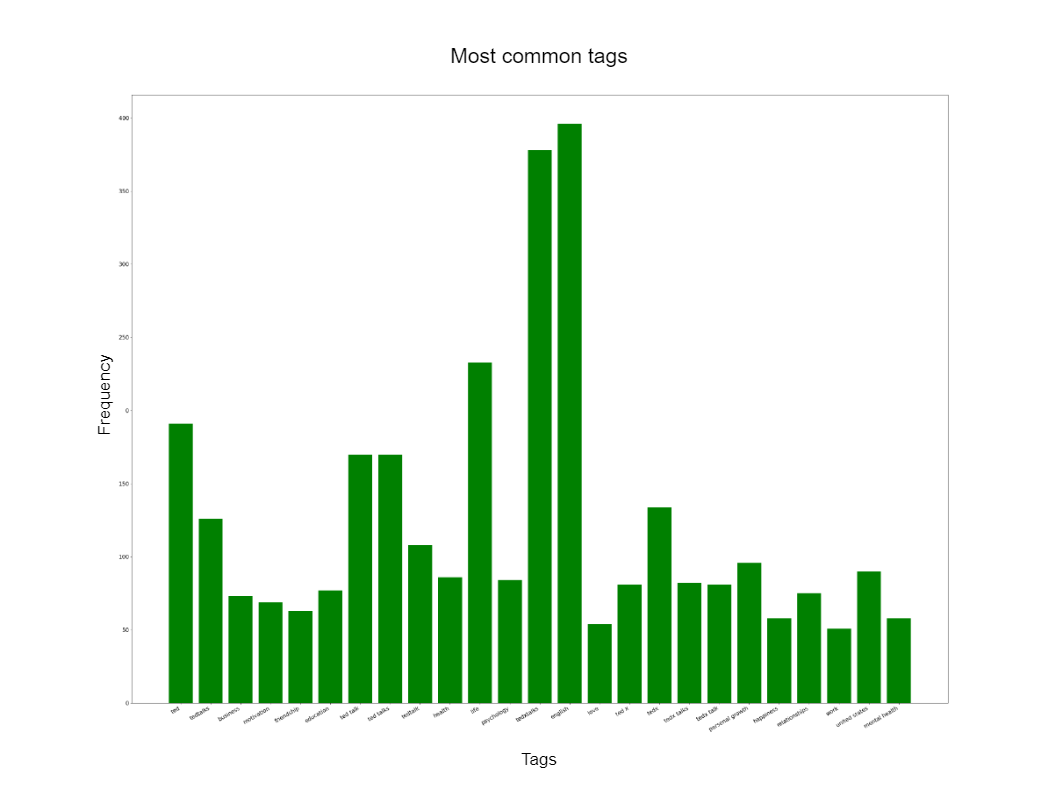} 
\caption{Frequency of youtube metadata tags gathered from TED videos in the dataset. From left to right: [ted, tedtalks, business, motivation, friendship, education, ted talk, ted talks, tedtalk, health, life, psychology, tedxtalks, English, love, ted x, tedx, tedx talk, personal growth, happiness, relationships, works, united states, mental health]}
\label{fig:dataset_examples}
\end{figure}

\begin{figure}[h]
\centering
\includegraphics[width=1\columnwidth]{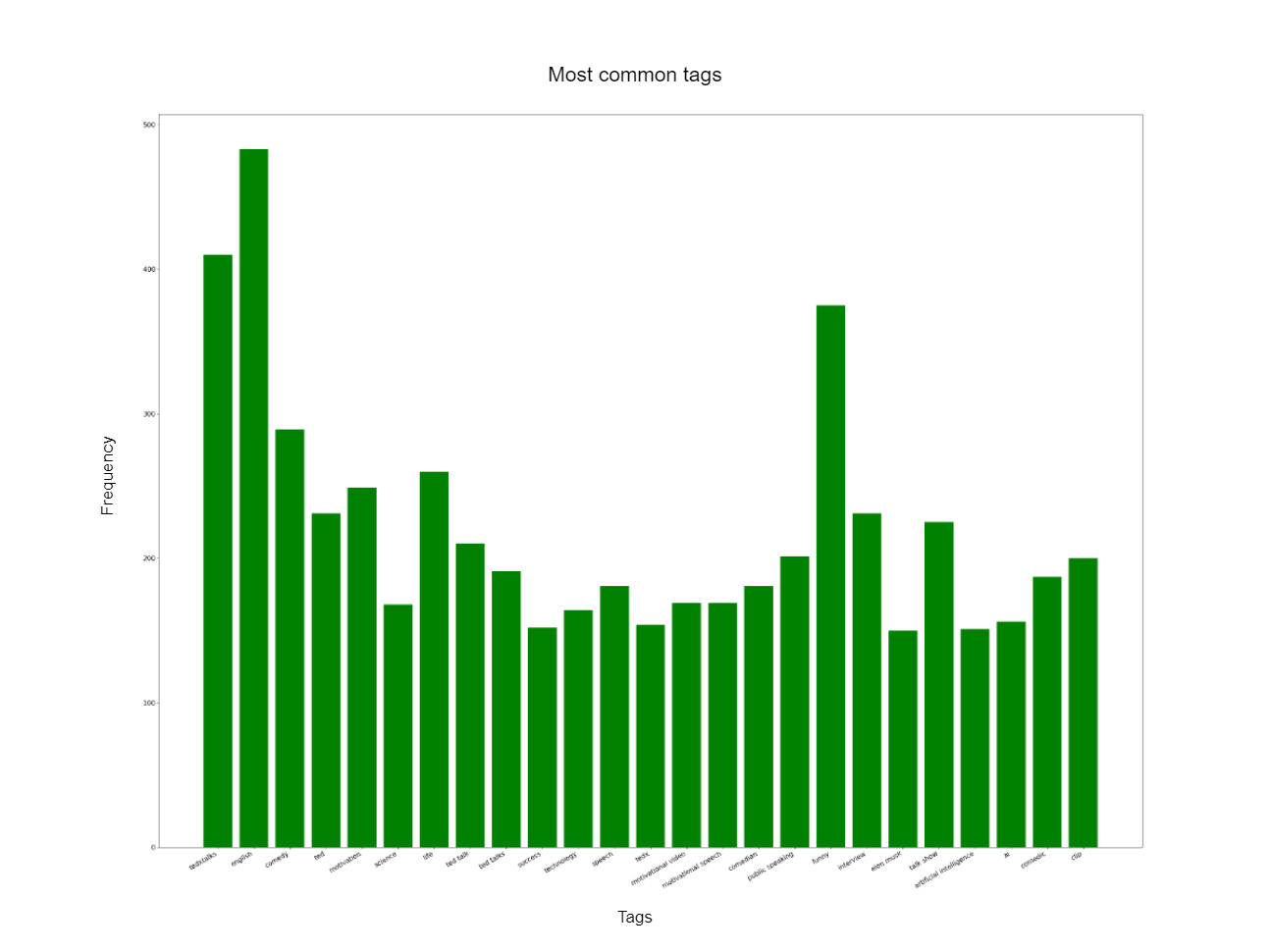} 
\caption{Frequency of youtube metadata tags gathered overall in the dataset. From left to right: [tedxtalks, English, comedy, ted, motivation, science, life, ted talk, ted talks, success, technology, speech, tedx, motivational video, comedian, public speaking, funny, interview, elon musk, talk show, Artificial intelligence, AI, comedic, clip]}
\label{fig:dataset_examples}
\end{figure}

\begin{figure}[h]
\centering
\includegraphics[width=1\columnwidth]{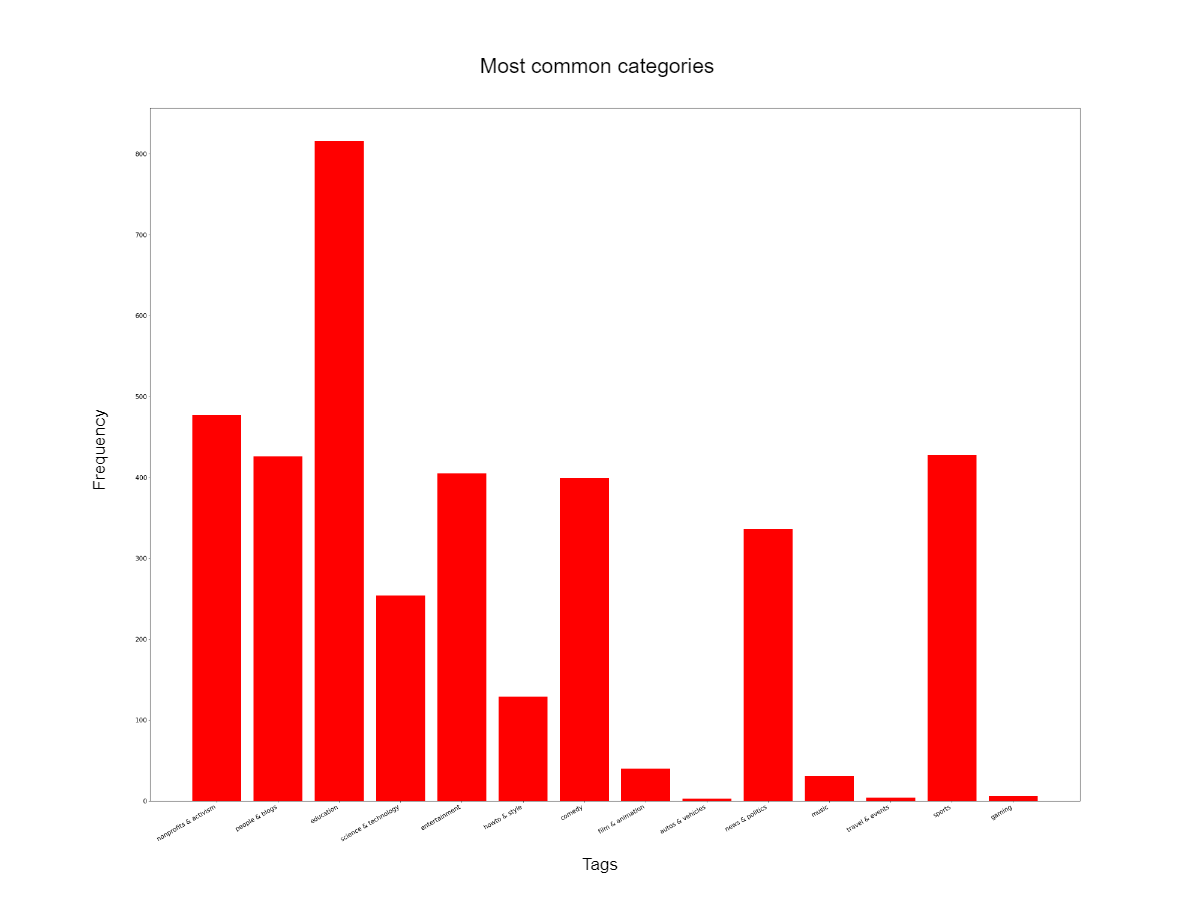} 
\caption{Frequency of youtube categories overall gathered in the dataset. From left to right: [nonprofits and activism, people and blogs, education, science and technology, entertainment, how-to and style, comedy, film and animation, autos and vehicles, news and politics, music, travel and events, sports, gaming]}
\label{fig:dataset_examples}
\end{figure}

\clearpage
\clearpage

\subsection{Compute for LBLM-AVA Training}

\begin{itemize} 
    \item \textbf{NVIDIA A100 Tensor Core GPUs}: Several A100 GPUs, each with 40 GB of memory, were used.
    \item \textbf{NVIDIA Titan RTX GPUs}: A few Titan RTX GPUs, each with 24 GB of memory, were also used.
    \item \textbf{NVIDIA GeForce RTX 2080 Ti GPUs}: Several 2080 Ti GPUs, each with 11 GB of memory, were included in the training setup.
\end{itemize}

\textbf{Training Details}:
\begin{itemize} 
    \item \textbf{Training Time}: Each complete training run took approximately 72 hours (using our entire dataset, and taking into account that our GPU cluster is a shared cluster), with the model trained for 50 epochs.
    \item \textbf{Batch Size and Optimizer}: A batch size of 32 samples per batch was used, and the AdamW optimizer with an initial learning rate of \(1 \times 10^{-4}\) was applied.
\end{itemize}

\textbf{Total Compute Estimation}:
\begin{itemize} 
    \item \textbf{GPU Hours}: With each run taking 72 hours on multiple GPUs, the total compute time was significant.
    \item \textbf{CPU Usage}: Each training run also utilized significant CPU resources, estimated at 144 CPU hours per run.
\end{itemize}

Preliminary experiments, model tuning, and failed attempts required approximately 500 additional GPU hours and 250 CPU hours.

\subsection{Experimental Setting and Details}

\begin{itemize}
    \item \textbf{Data Splits}: The dataset used in our experiments consists of 1,200 hours of multimodal data (video, audio, text). The data was divided into training, validation, and test sets with the following proportions: 70\% for training, 15\% for validation, and 15\% for testing.
    \item \textbf{Hyperparameters}: The hyperparameters were chosen randomly and fine-tuned through cross-validation. Key hyperparameters include:
    \begin{itemize}
        \item \textbf{Learning Rate}: The initial learning rate was set to \(1 \times 10^{-4}\) and was adjusted using a cosine annealing schedule.
        \item \textbf{Batch Size}: A batch size of 32 was chosen to balance between training efficiency and memory usage.
        \item \textbf{Optimizer}: The AdamW optimizer was used, known for its ability to handle sparse gradients and maintain stable training dynamics.
    \end{itemize}
    \item \textbf{Preprocessing Techniques}: Minor pre-processing techniques used in the dataset to train LBLM-AVA:
    \begin{itemize}
        \item \textbf{Video}: Frames were extracted at a rate of 30 frames per second and keypoints were detected using OpenPose and MediaPipe.
        \item \textbf{Text}: Text data was tokenized and embedded using pre-trained language models.
    \end{itemize}
    \item \textbf{Postprocessing Techniques}: Generated gestures were refined using temporal smoothing and alignment techniques to ensure realistic and coherent movements. These derivations and approaches are mentioned above.
\end{itemize}

\end{document}